%% file: paper.tex
\PassOptionsToPackage{table}{xcolor}
\documentclass[]{bytedance_seed}

\usepackage{amsmath}
\usepackage{amssymb}
\usepackage{array}
\usepackage{makecell}
\usepackage{algorithm}
\usepackage{algorithmic}
\usepackage{float}

\DeclareFontShape{T1}{bytesans}{b}{n}{<-> s * [1] seed/bytesans}{}

\title{Towards Full Pipeline FP8 Reinforcement Learning for LLMs}

\author[1,*]{Fanchao Chen}
\author[2]{Ziheng Jiang}
\author[2]{Ziyun Wei}
\author[2]{Zheng Zhong}
\author[2]{Du Li}
\author[2]{Chi Zhang}
\author[2]{Haibin Lin}
\author[1]{Shivaram Venkataraman}

\affiliation[1]{University of Wisconsin\textendash Madison}
\affiliation[2]{ByteDance Seed}
\contribution[*]{Work done during internship at ByteDance Seed}
\correspondence{Fanchao Chen at \email{fchen239@wisc.edu}}

\abstract{
Reinforcement learning (RL) has become a key technique for improving the reasoning and agentic abilities of large language models (LLMs). Although FP8 quantization can accelerate RL training, maintaining stability throughout an FP8 RL pipeline remains challenging. While previous works have focused on resolving train--inference mismatches using correction techniques like TIS, we reveal that full-pipeline FP8 RL still suffers from severe training instability, manifesting as anomalous mid-training entropy surges and garbled outputs. We trace this instability to a previously overlooked cause: compounded FP8 quantization noise distorts the importance ratio, disproportionately pushing negative-advantage tokens outside the trust region and erroneously zeroing out their gradients. As a result, pathological outputs are not properly penalized and accumulate over the course of training. To address this, we propose \textbf{Calibrated Clipping}, a dynamic method that aligns the FP8 clipping bounds with high-precision BF16 distributions by matching the lower-bound clipping quantile and rebalancing the upper bound accordingly. Extensive experiments across GRPO and DAPO algorithms, model scales from 8B to 32B, and multiple FP8 scaling granularities demonstrate that our approach successfully eliminates entropy surges and restores performance comparable to the BF16 baseline.
}

\begin{document}
\maketitle

\section{Introduction}
\label{sec:introduction}
\input{intro}

\section{Background}
\label{sec:background}
\input{backgrounds}

\section{Case Study}
\label{sec:case-study}
\input{study}

\section{Solution: Calibrated Clipping}
\label{sec:method}
\input{method}

\section{Evaluation}
\label{sec:evaluation}
\input{evaluation}
\FloatBarrier

\section{Related Work \& Discussion}
\label{sec:related-work}
\input{related}

\section{Conclusion}
\label{sec:conclusion}
\input{conclusion}

\bibliographystyle{plainnat}
\bibliography{main}

\clearpage
\beginappendix
\section{Additional Results and Experimental Details}
\label{sec:appendix}
\input{appendix}

\end{document}

%% file: intro.tex
The success of models such as DeepSeek-R1 \citep{guo2025deepseek} and o1 \citep{jaech2024openai} has demonstrated that reinforcement learning (RL) is capable of unlocking long-horizon reasoning and agentic capabilities in LLMs. As RL training continues to scale, improving its efficiency has become increasingly important. Recent advances in low-precision training, especially FP8, provide a promising direction for improving efficiency. While FP8 has already been successfully adopted in LLM pretraining \citep{liu2024deepseek} and inference \citep{qiu2026fp8}, recent efforts have begun extending it to RL pipelines.

However, enabling stable FP8 RL training remains challenging. Prior work \citep{yao2025flashrl, liu-li-2025-rl-collapse} has mainly focused on the mismatch between rollout and training introduced by FP8 quantization. Correction techniques such as truncated importance sampling (TIS, \cite{yao2025offpolicy}) have been proposed to mitigate this discrepancy, and more recent works \citep{xi2026jet, lmsys2025fp8rl} have explored unified FP8 pipelines—performing both rollout and training in FP8—to further reduce the precision gap and support on-policy training. Yet we find that training with an FP8 backend suffers from severe instability, manifesting as abrupt mid-training entropy surges (Figure \ref{fig:case}) that destabilize training and substantially degrade model performance.

Through empirical analysis and controlled case studies, we trace this instability to a previously overlooked cause: compounded quantization noise that perturbs the clipping mechanism in PPO-style surrogate objectives. Although the error on individual probabilities remains small, it is amplified in the importance ratio, distorting the effective trust region in quantized space. This effect is especially pronounced at the lower clipping bound, where negative-advantage tokens are disproportionately over-clipped, causing gradients that should suppress pathological generations to vanish prematurely. As a result, such pathological outputs are not properly penalized and accumulate over the course of training.

To address this, we propose \textbf{Calibrated Clipping}, a simple yet effective method for stable FP8 RL training. Our method periodically aligns the lower clipping bound using high-precision BF16 references and adjusts the upper bound to rebalance the contributions of positive and negative updates, restoring the intended trust-region semantics in quantized space with negligible overhead. Our main contributions are as follows:
\begin{itemize}
    \item We identify a previously overlooked failure mode in full-pipeline FP8 RL: compounded quantization noise in the importance ratio causes systematic over-clipping of tokens, triggering catastrophic mid-training entropy surges.
    \item We propose Calibrated Clipping, a method that restores trust-region semantics in quantized space by aligning the clipping bounds with a high-precision reference.
    \item Extensive experiments across GRPO and DAPO algorithms, model sizes from 8B to 32B, and multiple FP8 scaling granularities demonstrate that Calibrated Clipping consistently eliminates entropy surges, restores BF16-level performance, and can achieve up to 1.5$\times$ the BF16 training throughput.
\end{itemize}

%% file: backgrounds.tex
\subsection{Reinforcement Learning for LLMs}
\label{sec:rl-background}
Reinforcement learning (RL) has proven remarkably effective at eliciting complex reasoning and agentic behaviors in LLMs. Formally, given a prompt $x$ and a generated response $y$, the goal is to maximize the expected reward under the policy $\pi_\theta$. Modern RL pipelines adopt a proximal policy optimization (PPO) \citep{schulman2017proximal} surrogate objective for this goal:
\begin{equation}
    \mathcal{J}_{\text{PPO}}(\theta) = \mathbb{E}_{x \sim \mathcal{D},\, y \sim \pi_{\text{old}}(\cdot\mid x)} \left[ \sum_{t=1}^T \min\left(r_t(\theta) A_t, \operatorname{clip}\left(r_t(\theta), 1-\epsilon, 1+\epsilon\right) A_t\right) \right]
\end{equation}

Here, $A_t$ is the estimated advantage, typically computed via a learned critic model. Let $h_t=(x,y_{<t})$ denote the token context. The ratio $r_t(\theta) = \frac{\pi_\theta(y_t\mid h_t)}{\pi_{\text{old}}(y_t\mid h_t)}$ is the token-level likelihood ratio between the current and old policies used in the PPO surrogate objective. A more recent variant, GRPO \citep{shao2024deepseekmath}, removes the need for a learned critic model. Instead, for a group of $G$ responses, GRPO directly estimates the advantage using intra-group normalization $A_{i} = \frac{R_i - \operatorname{mean}(\{R_1, \dots, R_G\})}{\operatorname{std}(\{R_1, \dots, R_G\})}$.

The clipping mechanism in the PPO-style surrogate objective implicitly enforces a trust region between $\pi_{\theta}$ and $\pi_{\text{old}}$. 
For positive-advantage tokens, gradients are zeroed out when $r_t(\theta) \ge 1+\epsilon$; for negative-advantage tokens, gradients are zeroed out when $r_t(\theta) \le 1-\epsilon$, where typically $\epsilon=0.2$. Thus, the effective update contributions can be decomposed into two parts:
\begin{equation}
\begin{aligned}
\underbrace{\sum_{t:\,A_t>0}
\pi_{\text{old}}(y_t\mid h_t)\,
r_t(\theta)\,A_t\,
\mathbf{1}\{r_t(\theta)<1+\epsilon\}}_{\text{positive updates}}
\\
{}+
\underbrace{\sum_{t:\,A_t<0}
\pi_{\text{old}}(y_t\mid h_t)\,
r_t(\theta)\,A_t\,
\mathbf{1}\{r_t(\theta)>1-\epsilon\}}_{\text{negative updates}}.
\end{aligned}
\end{equation}
Positive updates encourage the model to increase the probabilities of advantageous tokens under the new policy, whereas negative updates tend to suppress the probabilities of disadvantageous tokens.

\subsection{FP8 Quantization}
The low-precision FP8 format accelerates training and inference but has limited representable range and precision. Thus, tensors are scaled before quantization \citep{fishman2024scaling}. Let $F_{\max}$ denote the largest finite value representable in the chosen FP8 format, and let $Q_{\mathrm{FP8}}$ denote rounding and clamping to the FP8 representable set. For a high-precision tensor $X$, the scaled FP8 tensor $\hat{X}$ and its dequantized approximation $\widetilde{X}$ are defined as
\begin{equation}
    S_X = \frac{\max(|X|)}{F_{\max}}, \qquad
    \hat{X} = Q_{\mathrm{FP8}}\left(\frac{X}{S_X}\right), \qquad
    \widetilde{X} = S_X\hat{X}.
\end{equation}

The granularity at which the maximum value $\max(|X|)$ is computed dictates the scaling strategy. There are three mainstream scaling strategies: \textbf{(1) tensorwise scaling}, where a single scaling factor $S_X$ is computed for the entire tensor $X$; \textbf{(2) rowwise scaling}, where independent scaling factors are calculated for each row of the tensor; and \textbf{(3) blockwise scaling}, where the tensor is partitioned into sub-blocks (e.g., $128 \times 128$ elements), with an independent $S_X$ assigned to each block \citep{liu2024deepseek}. These strategies span a spectrum of trade-offs between efficiency and numerical accuracy. Tensorwise scaling maximizes computational efficiency, but its coarse-grained scaling factor is vulnerable to outliers. Blockwise scaling offers the finest-grained adaptation and typically achieves the best numerical fidelity, but at the cost of additional scaling metadata and higher kernel complexity.

In practice, FP8 quantization has been widely adopted in LLM inference and pretraining. For example, DeepSeek-V3 \citep{liu2024deepseek} was pretrained in FP8 using blockwise scaling. Prior studies \citep{xi2026jet, qiu2026fp8} have also shown that FP8 inference can improve throughput by around 30\% without degrading downstream performance.

\subsection{Current Progress in FP8 RL for LLMs}
Following FP8's successful adoption in LLM pretraining and inference, migrating it to the RL stage is a natural progression. However, integrating FP8 into RL introduces unique challenges. 

Modern RL pipelines consist of two main components: rollout and training. They are usually decoupled in RL frameworks for better performance. Trajectories are sampled by high-throughput inference engines like vLLM \citep{kwon2023efficient} and SGLang \citep{zheng2024sglang}, while updates are computed in training backends like FSDP \citep{zhao2023pytorch} and Megatron-LM \citep{shoeybi2019megatron}.

Previous works \citep{yao2025flashrl, qiu2026fp8} have explored adopting FP8 in the rollout stage to accelerate generation. However, this design introduces a severe train--rollout mismatch: the behavior policy used by the sampler ($\textcolor{red}{\pi_{\text{sampler}}^{\mathrm{FP8}}}$) deviates from the learner's old policy ($\textcolor{blue}{\pi_{\text{old}}}$), even when both are instantiated from the same underlying model parameters. This discrepancy violates the on-policy assumption and can substantially degrade the stability of RL training. To mitigate this issue, truncated importance sampling (TIS, \cite{yao2025offpolicy,yao2025flashrl}) provides a truncated token-level correction:
\begin{equation}
\begin{aligned}
\mathcal{J}_{\text{TIS}}(\theta)
&= \mathbb{E}_{x \sim \mathcal{D},\, y \sim \textcolor{red}{\pi_{\text{sampler}}^{\text{FP8}}}(\cdot\mid x)}
\Bigg[ \sum_{t=1}^T
\underbrace{\min\left(
\frac{\textcolor{blue}{\pi_{\text{old}}}(y_t\mid h_t)}
{\textcolor{red}{\pi_{\text{sampler}}^{\text{FP8}}}(y_t\mid h_t)}, C\right)}_{w_t^{\text{TIS}}} \\
&\qquad \cdot
\min\left(r_t(\theta) A_t,\,
\operatorname{clip}\left(r_t(\theta), 1-\epsilon, 1+\epsilon\right) A_t\right)
\Bigg].
\end{aligned}
\end{equation}
Here, $C$ is the truncation threshold (typically set to 2).

Moving beyond using FP8 for rollout alone, more recent works \citep{xi2026jet, lmsys2025fp8rl} have explored a unified FP8 pipeline in which both rollout and training are performed in FP8. These studies showed that, compared with mixed-precision designs such as BF16 training with FP8 rollout, a unified FP8 pipeline can narrow the train--rollout log-probability gap, thereby improving overall training stability. However, as we show in the following sections, there are additional sources of instability that must be explicitly addressed to enable robust, full-pipeline FP8 RL training.

%% file: study.tex
\begin{figure}[t]
    \centering
    \includegraphics[width=\linewidth]{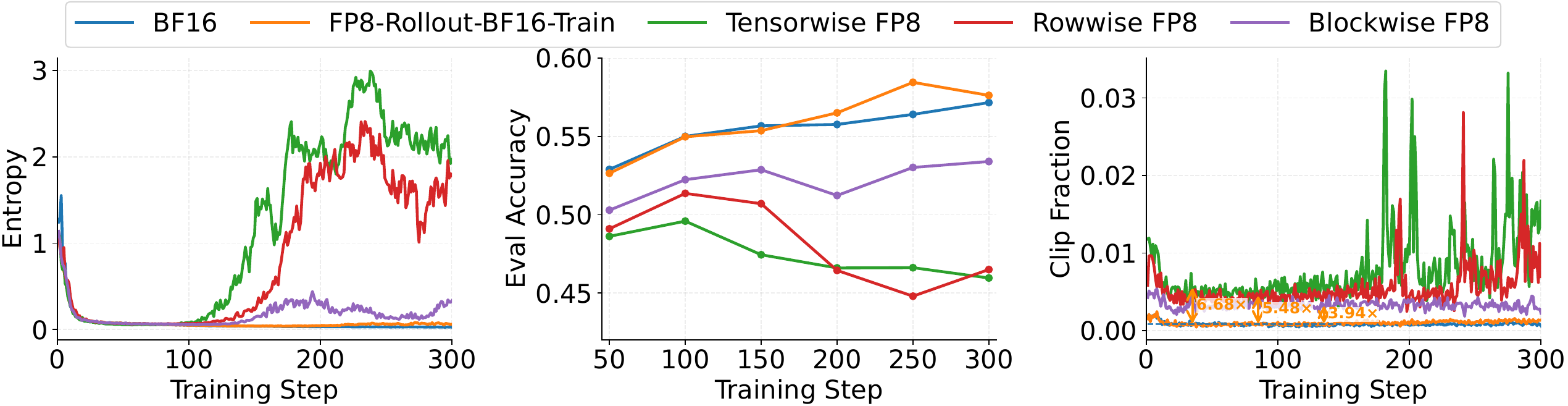}
    \caption{Comparison of (left) actor entropy, (middle) average evaluation accuracy on downstream tasks, and (right) PPO clipping fraction during the training of Qwen3-8B-Base under different settings.}
    \label{fig:case}
    \vspace{-1em}
\end{figure}

To investigate the stability of full-pipeline FP8 RL, we train Qwen3-8B-Base on the DeepScaleR dataset \citep{tan2025deepscaler} using GRPO with a 16K context length. We compare the standard BF16 baseline against configurations using FP8 rollout paired with four training backends: (1) BF16, (2) tensorwise FP8, (3) rowwise FP8, and (4) blockwise FP8. We adopt TIS by default in all experiments.

\textbf{The Phenomenon: Anomalous Entropy Surges.}
As shown in Figure \ref{fig:case}, we monitor the response entropy as an indicator of training stability. While the BF16 and FP8-Rollout-BF16-Train settings remain stable, the models trained with FP8 backends experience anomalous surges in entropy in the middle of training, typically after 100 steps. Consistent with our previous analysis, finer-grained scaling strategies mitigate quantization noise and lead to relatively milder surges; however, even with blockwise scaling, this pathological instability persists, resulting in a substantial performance disparity compared to the BF16 baseline.

\begin{figure}[htbp]
    \centering
    \begin{minipage}[t]{0.49\linewidth}
        \centering
        \includegraphics[height=0.58\linewidth]{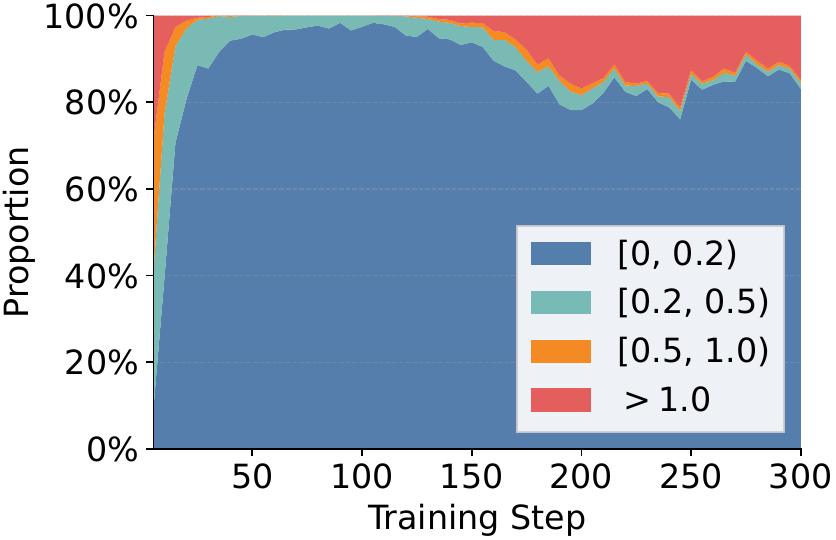}
        \caption{Composition of entropy across training steps in rowwise FP8 training.}
        \label{fig:proportion}
    \end{minipage}\hfill
    \begin{minipage}[t]{0.49\linewidth}
        \centering
        \includegraphics[height=0.58\linewidth]{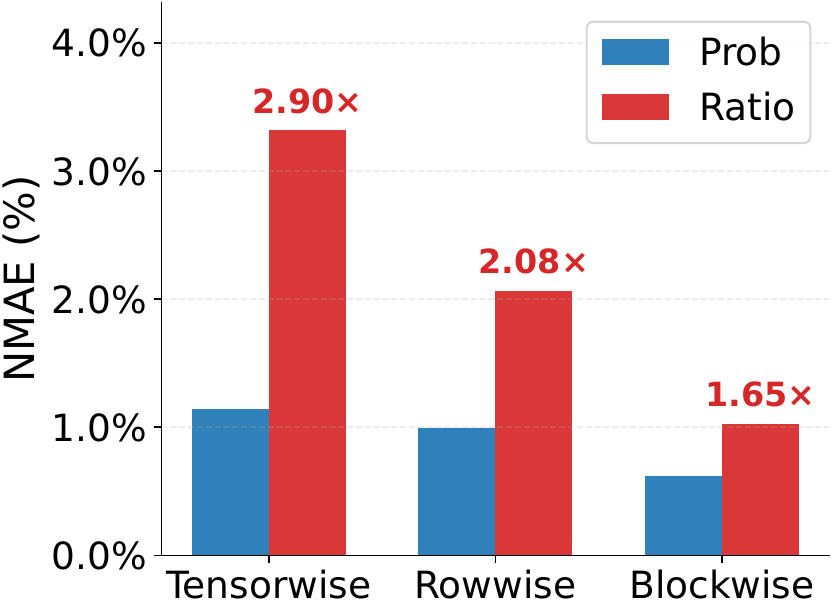}
        \caption{Normalized MAE of probabilities and importance ratios from the FP8 model relative to BF16 references.}
        \label{fig:nmae}
    \end{minipage}
\end{figure}

Since this instability persists across all scaling granularities, we adopt rowwise FP8 as the representative configuration to diagnose the underlying mechanics. As illustrated in Figure \ref{fig:proportion}, we analyze the composition of response entropies by partitioning them into four discrete bins: $[0, 0.2), [0.2, 0.5), [0.5, 1.0),$ and $[1.0, \infty)$. Our analysis reveals that the global entropy surge is non-uniform; while the vast majority of responses maintain healthy entropy levels ($<0.5$), the surge is driven by a small but rapidly proliferating fraction of samples exhibiting pathological entropy ($> 1.0$), which spikes sharply in frequency after 100 steps.

After inspecting these high-entropy responses, we find that they generally contain a large proportion of garbled, fragmented, or nonsensical tokens, signifying a collapse in the model's decoding state (Figure \ref{fig:response}). Further analysis of the associated advantages shows that only 6.64\% of them yield positive advantages. With the policy optimization described in Section~\ref{sec:rl-background}, the remaining majority should be penalized to suppress their occurrence. Yet we observe that their prevalence continues to grow.

\textbf{The Cause: Compounded Noise and Over-Clipping.}
We trace this counterintuitive failure to the interaction between FP8 quantization noise and the PPO clipping mechanism. 

In full-pipeline FP8 training, the probabilities for both $\pi_{\theta}$ and $\pi_{\text{old}}$ are induced by FP8-quantized forward passes. Using BF16 shadow passes for diagnosis, we measured the normalized mean absolute error (NMAE) of these FP8-induced probabilities relative to high-precision references. As illustrated in Figure \ref{fig:nmae}, while the quantization error for individual probabilities is marginal, the ratio operation significantly compounds this noise, enlarging the discrepancy by $1.7\times$ to $2.9\times$.

This compounded noise inflates the frequency at which tokens hit the trust region boundaries, a phenomenon we define as \textit{over-clipping}. A token is lower-bound over-clipped if $A_t<0$, $r_t^{\mathrm{BF16}}>1-\epsilon$, and $r_t^{\mathrm{FP8}}\le 1-\epsilon$; analogously, it is upper-bound over-clipped if $A_t>0$, $r_t^{\mathrm{BF16}}<1+\epsilon$, and $r_t^{\mathrm{FP8}}\ge 1+\epsilon$. The PPO clipping fractions shown in Figure \ref{fig:case} provide empirical evidence for this phenomenon. Even with fine-grained blockwise scaling, the PPO clipping fraction remains nearly $4\times$ higher than in the BF16 baseline. 

\begin{figure}[htbp]
    \centering
    \begin{minipage}[t]{0.37\linewidth}
        \centering
        \includegraphics[width=\linewidth]{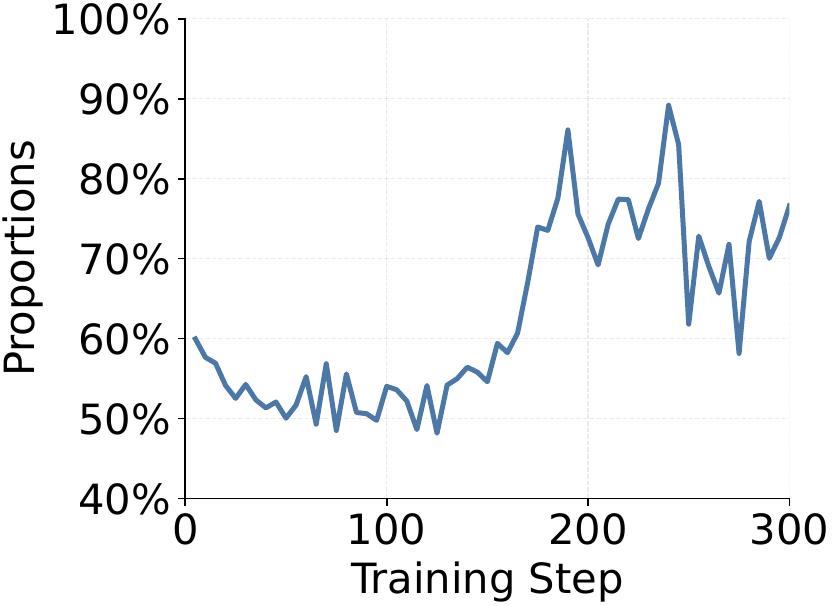}
    \end{minipage}\hspace{0.04\linewidth}%
    \begin{minipage}[t]{0.37\linewidth}
        \centering
        \includegraphics[width=\linewidth]{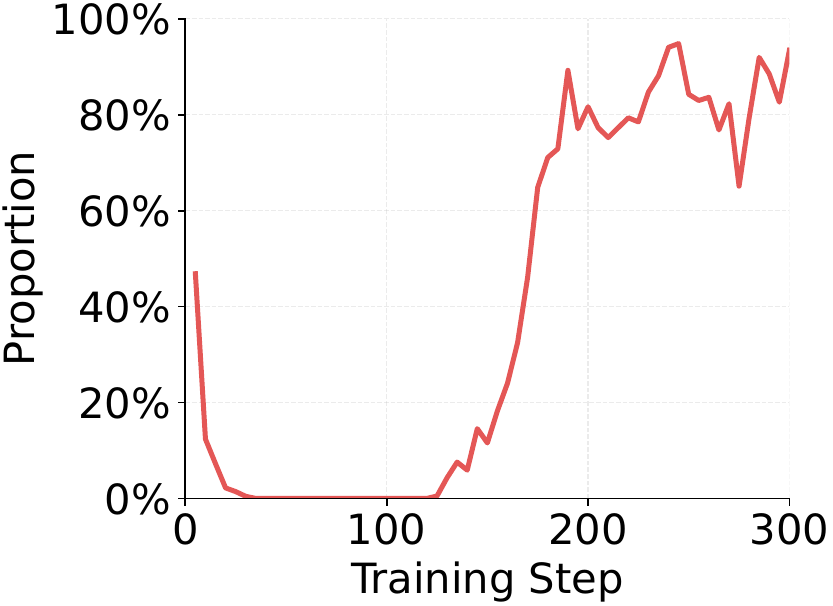}
    \end{minipage}
    \caption{Analysis of over-clipping: (Left) Proportion of over-clipping occurrences at the lower bound ($1-\epsilon$). (Right) Proportion of lower-bound over-clipped tokens originating from high-entropy responses.}
    \label{fig:case_trend}
\end{figure}

Our evaluation of these over-clipped tokens reveals two asymmetric trends (Figure \ref{fig:case_trend}):
\begin{itemize}
    \item During the later stages of training, approximately 70\% of over-clipping occurs at the lower bound ($1-\epsilon$). 
    \item Nearly 90\% of these lower-bound over-clipped tokens originate from the high-entropy responses identified in Figure \ref{fig:proportion}.
\end{itemize}

We identify the failure loop as follows: when the model generates garbled responses with negative advantages, the policy optimization attempts to penalize them. However, the compounded FP8 noise frequently pushes the importance ratios below the $1-\epsilon$ threshold, falsely triggering the clipping mechanism and zeroing out the gradients. This premature gradient vanishing prevents the model from unlearning pathological behaviors, allowing garbled outputs to proliferate and eventually destabilize training.

\begin{figure}[H]
    \centering
    \includegraphics[width=\linewidth]{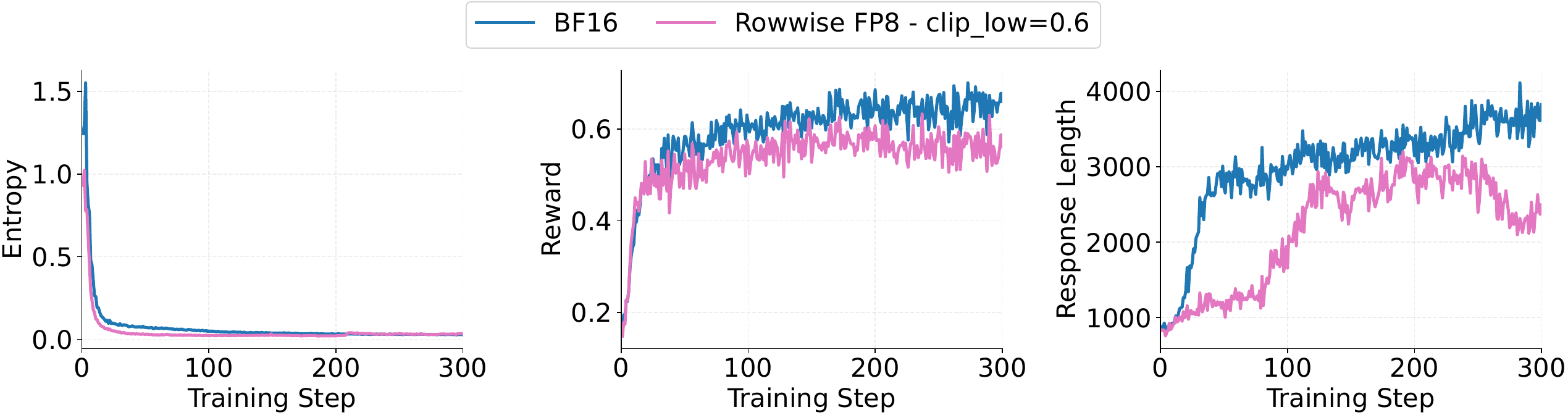}
    \caption{Relaxing the lower bound in rowwise FP8 training prevents the entropy surge (left) but leads to premature entropy saturation, lower training rewards (middle), and shorter responses (right).}
    \label{fig:low}
    \vspace{-1em}
\end{figure}

\textbf{Empirical Verification and Implications.}
To verify our analysis, we reran the rowwise FP8 experiment, relaxing the lower clipping bound from 0.8 to 0.6. As shown in Figure \ref{fig:low}, this simple adjustment successfully prevented the catastrophic mid-training entropy surge.

However, it introduces a new problem. While a looser lower bound restores the ability to penalize garbled responses, it results in excessive negative updates that cause the entropy to saturate prematurely. This ``over-penalization'' also hurts model performance; as shown in Figure \ref{fig:low}, the average reward and response length remain consistently lower than the BF16 baseline, suggesting that the model has become overly conservative. These results demonstrate that relaxing the lower bound in isolation is insufficient. We need a principled approach to rebalance the trust region.

%% file: method.tex
Integrating FP8 into the training backend alters the trust region landscape. Under the FP8 regime, probabilities are computed via quantized forward passes, yielding the importance ratio: $r_t^{\text{FP8}} = \tilde{Q}(r_t; \theta, \theta_{\text{old}})$, where $\tilde{Q}(\cdot)$ captures the compounded, nonlinear effect of FP8 quantization. To preserve the original trust region semantics, the ideal clipping bounds $[L, H]$ should satisfy $L = \tilde{Q}(1-\epsilon; \theta, \theta_{\text{old}}), \quad H = \tilde{Q}(1+\epsilon; \theta, \theta_{\text{old}})$.

In practice, the exact formulation of $\tilde{Q}(\cdot)$ is intractable. As discussed in Section~\ref{sec:case-study}, the lower boundary is particularly important: garbled responses predominantly carry negative advantages, and FP8-induced ratio perturbations increase the likelihood that their ratios cross this boundary. Thus, we propose \textbf{Calibrated Clipping}, a method that dynamically reconstructs the trust region through two stages: 

\textbf{1. Lower-Bound Alignment.} We first align the clipping quantile of the FP8 distribution with that of BF16 for negative-advantage tokens. The goal is to ensure that the proportion of negative-advantage tokens clipped at the lower bound in the FP8 domain matches that in BF16. 

For a given training step, let $F_{\text{BF16},-}(r)$ and $F_{\text{FP8},-}(r)$ denote the cumulative distribution functions (CDFs) of the importance ratios under BF16 and FP8, respectively, restricted to tokens with $A_t<0$. We first determine the target clipping quantile $\alpha_{\text{low}}$ from the reference BF16 distribution: $\alpha_{\text{low}} = F_{\text{BF16},-}(1-\epsilon)$. We then derive the calibrated lower bound $L$ by applying the inverse CDF to the FP8 distribution: $L = F_{\text{FP8},-}^{-1}(\alpha_{\text{low}})$. Replacing $1-\epsilon$ with the calibrated quantile $L$ equalizes the number of lower-clipped tokens under the two precisions and effectively prevents over-clipping.

\textbf{2. Upper-Bound Rebalancing.} To mitigate over-penalization from a relaxed lower bound, we rebalance the positive-to-negative update ratio. In Section~\ref{sec:rl-background}, we showed that the effective contribution to the surrogate objective can be decomposed into positive and negative parts. For clipping bounds $[c_{\text{low}}, c_{\text{high}}]$ and TIS weights $w_t^{\text{TIS}}$, the ratio between the positive and negative parts can be computed as:
\begin{equation}
\rho = \frac{
\left| \sum_{A_t>0} w_t^{\text{TIS}} \pi_{\text{sampler}}(y_t\mid h_t) r_t(\theta) A_t \mathbf{1}\{r_t(\theta) < c_{\text{high}}\} \right|
}{
\left| \sum_{A_t<0} w_t^{\text{TIS}} \pi_{\text{sampler}}(y_t\mid h_t) r_t(\theta) A_t \mathbf{1}\{r_t(\theta) > c_{\text{low}}\} \right|
}
\end{equation}

\begin{figure}[htbp]
    \centering
    \includegraphics[width=0.36\linewidth]{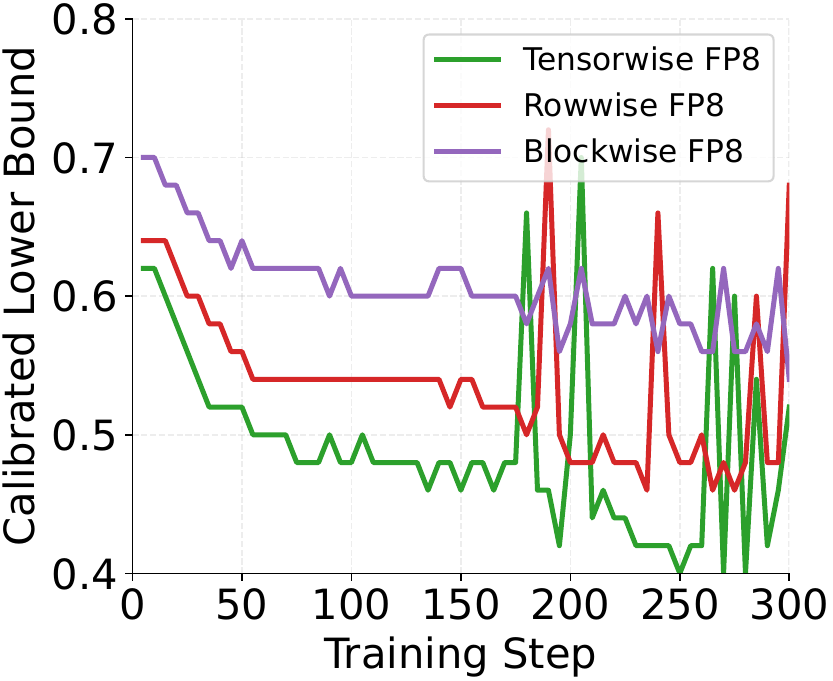}
    \caption{Trajectory of the calibrated lower bound.}
    \label{fig:clip_low}
\end{figure}

where $w_t^{\text{TIS}} = \min\left( \frac{\pi_{\text{old}}(y_t\mid h_t)}{\pi_{\text{sampler}}(y_t\mid h_t)}, C \right)$. We compute the reference ratio $\rho_{\text{BF16}}$ from the BF16 distribution. Then, given the calibrated lower bound $L$, we solve for a new upper bound $H$ such that the resulting ratio $\rho_{\text{FP8}}$ matches $\rho_{\text{BF16}}$. 

\textbf{Periodic Recalibration.}
Both stages of our method require BF16 reference results. At each recalibration step, we perform two forward-only BF16 shadow passes over the same training batch, evaluating the current policy $\pi_\theta^{\mathrm{BF16}}$ and the old policy $\pi_{\mathrm{old}}^{\mathrm{BF16}}$ using their master weights. We then compute the reference importance ratios $r_t^{\mathrm{BF16}}$. However, performing these additional forward passes at every training step would negate the performance gains of FP8 training.

Fortunately, we find that the calibrated trust region does not fluctuate drastically when training is stable. Figure \ref{fig:clip_low} illustrates the trajectory of the calibrated lower bound derived from our case study in Section~\ref{sec:case-study}. Mirroring the evolution of actor entropy, the calibrated bound remains stable after the initial phase of training. Notable inter-step deviations are only observed in the later stages during entropy surges. Leveraging this stability, we can amortize the overhead by periodically recalculating the clipping bounds rather than performing per-step alignment.

Our method is formalized in Algorithm \ref{alg:calibclip}. We define the search intervals as $\mathcal{I}_L = [0.5, 0.9]$ for the lower bound and $\mathcal{I}_H = [1.2, 2.0]$ for the upper bound, employing a search step size of $\Delta = 0.02$. We define $\operatorname{SmoothUpdate}(b,b^\star,\delta)=b+\operatorname{clip}\left((b^\star-b)/2,-\delta,\delta\right)$, with maximum update magnitudes $\delta_L = 0.05$ and $\delta_H = 0.1$. This prevents abrupt shifts in the trust region, maintaining a smooth evolution. We initialize the clipping bounds at $[0.6, 1.8]$ for faster convergence and perform updates every 20 training steps. While these hyperparameters were not explicitly tuned, they demonstrate robust performance across all evaluated settings. We evaluate the sensitivity to the update interval and clipping-bound initialization in Appendix~\ref{sec:sensitivity-appendix}.

%% file: evaluation.tex
\begin{algorithm}[t]
\caption{Calibrated Clipping}
\label{alg:calibclip}
\begin{algorithmic}[1]
\item[] \textbf{Input:} Batch advantages $A$, TIS weights $w^{\text{TIS}}$, importance ratios $r^{\text{BF16}}$ and $r^{\text{FP8}}$, previous bounds $L_{\text{old}}$ and $H_{\text{old}}$, reference bounds $c_{\text{low}}^{\text{ref}}$ and $c_{\text{high}}^{\text{ref}}$, search intervals $\mathcal{I}_L$ and $\mathcal{I}_H$, search step $\Delta$, and update constraints $\delta_L$ and $\delta_H$.

\STATE \textbf{Procedure} Derive high-precision reference statistics
\STATE \quad Compute the reference lower quantile over tokens with $A_t<0$: $\alpha_{\text{low}} = F_{\text{BF16},-}(c_{\text{low}}^{\text{ref}})$;
\STATE \quad Compute reference update ratio $\rho_{\text{BF16}}$ using $A$, $w^{\text{TIS}}$, $r^{\text{BF16}}$, and bounds $[c_{\text{low}}^{\text{ref}}, c_{\text{high}}^{\text{ref}}]$;

\STATE \textbf{Procedure} Search for optimal FP8 clipping bounds
\STATE \quad Search $L^* \in \mathcal{I}_L$ with step size $\Delta$ to minimize $|F_{\text{FP8},-}(L^*) - \alpha_{\text{low}}|$;
\STATE \quad Search $H^* \in \mathcal{I}_H$ with step size $\Delta$ to minimize $|\rho_{\text{FP8}}(L^*, H^*) - \rho_{\text{BF16}}|$;

\STATE \textbf{Procedure} Bound smoothing and update
\STATE \quad $L_{\text{new}} = \operatorname{SmoothUpdate}(L_{\text{old}}, L^*, \delta_L)$;
\STATE \quad $H_{\text{new}} = \operatorname{SmoothUpdate}(H_{\text{old}}, H^*, \delta_H)$;
\STATE \textbf{return} $L_{\text{new}}, H_{\text{new}}$

\end{algorithmic}
\end{algorithm}

All experiments use VeRL \citep{sheng2025hybridflow} as the RL framework, vLLM \citep{kwon2023efficient} as the inference engine, and TorchAO \citep{or2025torchao} as the FP8 training engine. We integrate the patch provided by FlashRL \citep{yao2025flashrl} for FP8 rollout. To compare training stability across settings, all reported metrics were collected over the final 50\% of training steps. Experiment configurations and additional training metrics—such as actor entropy, rewards, response lengths, clipping fractions, and calibrated clipping bounds—are detailed in the Appendix.

\subsection{Results of GRPO Experiments}
\label{sec:grpo-results}
We train Qwen3-8B-Base and Qwen2.5-32B with GRPO on the DeepScaleR dataset \citep{tan2025deepscaler}. We evaluate the models on eight widely used reasoning benchmarks, including AIME24/25 \citep{aops2024aime}, OlympiadBench \citep{he2024olympiadbench}, Gaokao \citep{zhang2023evaluating}, MATH-500 \citep{hendrycks2021measuring,lightman2023lets}, Minerva Math \citep{lewkowycz2022solving}, and AMC23/24 \citep{aops2024amc}. We run the experiments for 500 steps and evaluate on these benchmarks every 25 steps. We report the performance of the checkpoint that achieves the best average score. For clipping calibration, we set $c_{\text{low}}^{\text{ref}} = 0.8$ and $c_{\text{high}}^{\text{ref}} = 1.24$ to prevent entropy saturation.

\begin{table}[H]
\centering
\small
\setlength{\tabcolsep}{2.2pt}
\renewcommand{\arraystretch}{1.1}

\definecolor{lightgrayrow}{gray}{0.94}
\definecolor{darkgreen}{RGB}{0,150,0}
\definecolor{darkred}{RGB}{200,0,0}

\begin{tabular}{@{}>{\centering\arraybackslash}m{2.05cm} >{\centering\arraybackslash}m{1.05cm} c c c c c c >{\centering\arraybackslash}m{2.6cm}@{}}
\toprule
Train & Rollout & AIME & AMC & MATH & Gaokao & Minerva & Olymp. & Average \\
\midrule
\multicolumn{9}{l}{\textit{Qwen3-8B-Base}} \\
\midrule
BF16 & BF16 & 31.3/26.0 & 76.9/62.8 & 89.6 & 71.9 & 44.0 & 58.1 & 57.6 \\
BF16 & FP8 & 36.9/25.2 & 73.8/60.6 & 89.6 & 75.8 & 44.8 & 59.4 & 58.2 \\
\midrule
Tensorwise FP8 & FP8 & 11.5/14.8 & 56.9/45.0 & 82.9 & 68.7 & 42.7 & 46.3 & 46.1 \\
\rowcolor{lightgrayrow} w/ Calibrated Clipping & FP8 & 29.0/22.5 & 73.1/57.2 & 88.6 & 74.5 & 44.6 & 57.6 & \makecell[c]{55.9 \\ ({\color{darkred}-1.7},{\color{darkred}-2.3},{\color{darkgreen}+9.8})} \\
\midrule
Rowwise FP8 & FP8 & 15.0/17.3 & 57.5/43.3 & 83.4 & 68.1 & 42.2 & 49.0 & 47.0 \\
\rowcolor{lightgrayrow} w/ Calibrated Clipping & FP8 & 31.3/21.3 & 76.3/58.9 & 89.3 & 73.6 & 43.4 & 58.3 & \makecell[c]{56.5 \\ ({\color{darkred}-1.1},{\color{darkred}-1.7},{\color{darkgreen}+9.5})} \\
\midrule
Blockwise FP8 & FP8 & 25.8/21.9 & 75.0/50.0 & 88.1 & 72.1 & 45.0 & 54.6 & 54.1 \\
\rowcolor{lightgrayrow} w/ Calibrated Clipping & FP8 & 33.5/27.1 & 80.6/62.8 & 90.7 & 75.6 & 41.3 & 57.3 & \makecell[c]{58.6 \\ ({\color{darkgreen}+1.0},{\color{darkgreen}+0.4},{\color{darkgreen}+4.5})} \\
\midrule
\multicolumn{9}{l}{\textit{Qwen2.5-32B}} \\
\midrule
BF16 & BF16 & 27.5/16.0 & 71.3/50.0 & 85.4 & 71.6 & 41.9 & 51.9 & 51.9 \\
BF16 & FP8 & 21.5/17.3 & 71.9/52.8 & 85.5 & 70.4 & 41.4 & 49.0 & 51.2 \\
\midrule
Tensorwise FP8 & FP8 & 24.4/16.9 & 65.0/40.0 & 84.6 & 71.1 & 42.1 & 48.5 & 49.1 \\
\rowcolor{lightgrayrow} w/ Calibrated Clipping & FP8 & 25.0/16.3 & 73.8/48.9 & 83.7 & 69.4 & 41.7 & 50.0 & \makecell[c]{51.1 \\ ({\color{darkred}-0.8},{\color{darkred}-0.1},{\color{darkgreen}+2.0})} \\
\midrule
Rowwise FP8 & FP8 & 20.4/15.2 & 66.3/43.9 & 84.7 & 71.2 & 43.1 & 47.3 & 49.0 \\
\rowcolor{lightgrayrow} w/ Calibrated Clipping & FP8 & 24.6/17.9 & 70.0/51.7 & 86.5 & 71.6 & 42.1 & 49.1 & \makecell[c]{51.7 \\ ({\color{darkred}-0.2},{\color{darkgreen}+0.5},{\color{darkgreen}+2.7})} \\
\midrule
Blockwise FP8 & FP8 & 26.9/17.5 & 71.3/46.1 & 85.1 & 72.8 & 40.3 & 46.4 & 50.8 \\
\rowcolor{lightgrayrow} w/ Calibrated Clipping & FP8 & 25.6/19.8 & 70.0/52.8 & 86.5 & 73.2 & 45.3 & 54.1 & \makecell[c]{53.4 \\ ({\color{darkgreen}+1.5},{\color{darkgreen}+2.2},{\color{darkgreen}+2.6})} \\
\bottomrule
\end{tabular}

\caption{Performance comparison for \textbf{Qwen3-8B-Base} and \textbf{Qwen2.5-32B} under different training precisions and rollout configurations. AIME, AMC, and MATH denote AIME 2024/2025, AMC 2023/2024, and MATH-500, respectively. The three signed values in parentheses in the Average column denote performance changes relative to (1) the BF16 baseline, (2) FP8 rollout with BF16 training, and (3) the corresponding vanilla FP8 training baseline. Red indicates a decrease and green an increase.}
\label{tab:grpo_result}
\vspace{-1em}
\end{table}

As shown in Table~\ref{tab:grpo_result}, vanilla FP8 training causes significant performance drops for Qwen3-8B-Base, with average scores declining by 11.5, 10.6, and 3.5 points for tensorwise, rowwise, and blockwise scaling, respectively. Calibrated Clipping substantially closes these gaps across all granularities, with blockwise FP8 reaching an average score comparable to the BF16 baseline (58.6 vs. 57.6). For Qwen2.5-32B, pathological entropy surges are less pronounced—attributable to its significantly shorter responses—yet Calibrated Clipping still yields consistent gains, with blockwise FP8 likewise reaching a comparable average score (53.4 vs. 51.9). We also report the Qwen3-8B-Base results on the Eurus coding dataset in Appendix~\ref{sec:coding-appendix}.

\subsection{Results of DAPO Experiments}
\label{sec:dapo-results}
We train Qwen3-14B-Base with the DAPO algorithm on the 17K dataset from the original paper \citep{yu2025dapo}. We train for 200 steps and evaluate Avg@32 accuracy on AIME24 every 10 steps. We set $c_{\text{low}}^{\text{ref}} = 0.8$ and $c_{\text{high}}^{\text{ref}} = 1.28$ to align with the DAPO configuration.

The evaluation results are summarized in Table~\ref{tab:dapo_results}. In contrast to the GRPO experiments, the higher upper clipping bound in DAPO prevents entropy from saturating; instead, even the high-precision baseline exhibits a gradual entropy increase throughout training, as shown in Figure \ref{fig:14b_dapo}. This inherent property makes vanilla FP8 training more prone to triggering pathological entropy surges. By implementing Calibrated Clipping, we successfully mirror the baseline's entropy evolution while recovering a substantial portion of the performance lost under vanilla FP8 training across all scaling granularities. 

\begin{table}[H]
\centering
\small
\setlength{\tabcolsep}{4pt}
\renewcommand{\arraystretch}{1.1}

\definecolor{lightgrayrow}{gray}{0.94}
\definecolor{darkgreen}{RGB}{0,150,0}
\definecolor{darkred}{RGB}{200,0,0}

\begin{tabular}{>{\centering\arraybackslash}m{4.5cm} >{\centering\arraybackslash}m{1.8cm} >{\centering\arraybackslash}m{5.0cm}}
\toprule
Train & Rollout & AIME24 Avg@32 \\
\midrule
BF16 & BF16 & 50.9 \\
BF16 & FP8  & 47.4 \\
\midrule
Tensorwise FP8 & FP8 & 35.7 \\
\rowcolor{lightgrayrow} w/ Calibrated Clipping & FP8 & \mbox{47.9 ({\color{darkred}-3.0}, {\color{darkgreen}+0.5}, {\color{darkgreen}+12.2})} \\
\midrule
Rowwise FP8 & FP8 & 38.1 \\
\rowcolor{lightgrayrow} w/ Calibrated Clipping & FP8 & \mbox{46.5 ({\color{darkred}-4.4}, {\color{darkred}-0.9}, {\color{darkgreen}+8.4})} \\
\midrule
Blockwise FP8 & FP8 & 41.6 \\
\rowcolor{lightgrayrow} w/ Calibrated Clipping & FP8 & \mbox{47.4 ({\color{darkred}-3.5}, 0.0, {\color{darkgreen}+5.8})} \\
\bottomrule
\end{tabular}

\caption{Performance evaluation of \textbf{Qwen3-14B-Base} on the AIME24 benchmark using the DAPO algorithm. As in Table~\ref{tab:grpo_result}, the red/green values in parentheses denote the corresponding performance changes.}
\label{tab:dapo_results}
\end{table}

\begin{figure}[t]
    \centering
    \includegraphics[width=\linewidth]{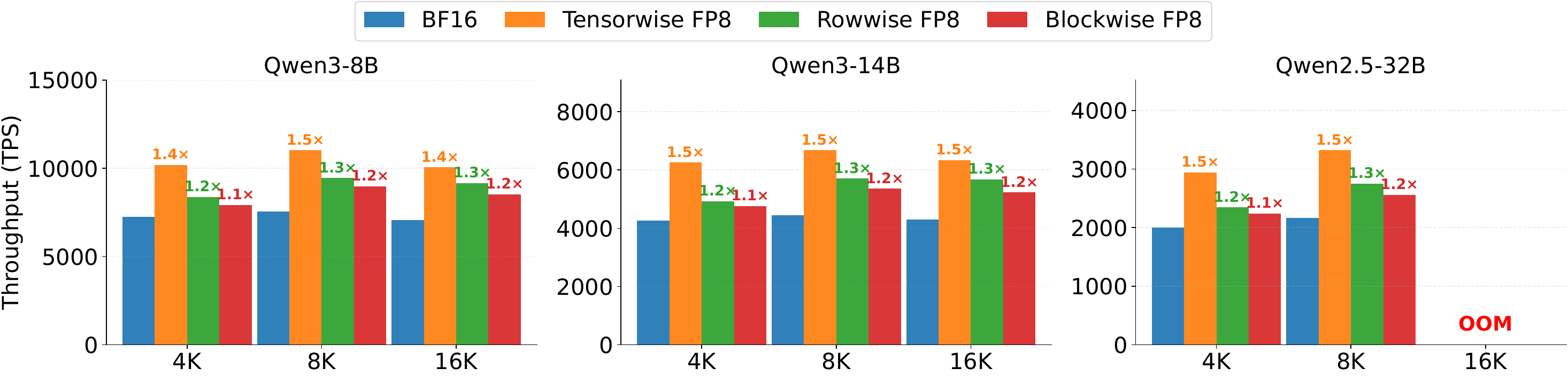}
    \caption{Throughput (tokens per second) of FP8 training compared with the BF16 baseline across model sizes (8B, 14B, 32B) and sequence lengths (4K, 8K, 16K). The offline TorchAO benchmark isolates training-phase throughput and excludes periodic BF16 reference passes.}
    \label{fig:perf}
    \vspace{-1em}
\end{figure}

\subsection{Performance Gains}
We focus on performance gains within the training phase enabled by full-pipeline FP8 training, benchmarking across 8B, 14B, and 32B models at sequence lengths of 4K, 8K, and 16K. As shown in Figure \ref{fig:perf}, coarser scaling granularities generally yield greater throughput improvements. Tensorwise scaling typically achieves up to 1.5$\times$ the BF16 training throughput, while blockwise scaling can also provide a 10\%--20\% improvement. These training gains, together with the approximately 30\% generation speedup from FP8 rollout reported by prior work \citep{yao2025flashrl}, enable substantial full-pipeline efficiency improvements.

%% file: related.tex
\textbf{Related RL Algorithms and Clipping Mechanisms.}
Several works have sought to improve RL stability by redesigning the clipping mechanism.
DAPO \citep{yu2025dapo} proposes asymmetrically relaxing the upper bound to encourage exploration.
GSPO \citep{zheng2025group} reconceives the trust region at the sequence level,
applying clipping over entire responses rather than individual tokens.
Beyond importance-ratio clipping, M2PO \citep{zheng2025prosperity} identifies extreme updates
using a batch-level second-moment statistic over log-importance ratios,
while DPPO \citep{qi2026rethinking} directly determines the trust region from policy divergence.
Our analysis focuses on token-level clipping in GRPO and DAPO.
Sequence-level aggregation may reduce sensitivity to individual token-ratio errors,
but precision-induced distortion may still affect the aggregated sequence ratio;
extending calibration to sequence-level clipping remains future work.

\citet{chen2026exploration} connect clipping bias to policy-entropy reduction under spurious rewards,
while \citet{karaman2026dispo} decouple importance-weight clipping for correct and incorrect responses
to balance exploration and training stability.
These studies analyze entropy control under reward or off-policy effects;
our work instead focuses on false lower-bound clipping induced by FP8 ratio distortion.

CISPO \citep{chen2025minimax} is closely related to our work.
It addresses gradient zeroing from hard clipping by retaining gradient contributions from clipped tokens
and clipping only the sampling weights.
However, CISPO was not designed for FP8 RL.
As shown in Appendix~\ref{sec:cispo-appendix}, applying CISPO to rowwise FP8 training
leads to training collapse after 300 steps,
indicating that the clipping boundaries still need to be carefully calibrated.
BAPO \citep{xi2025bapo} proposes dynamically adjusting the clipping bounds
to rebalance positive and negative contributions.
While BAPO adjusts its bounds to satisfy a predefined threshold on the positive contribution to the policy loss,
our Calibrated Clipping derives the lower bound and the target ratio directly from a high-precision reference
and solves for the upper bound accordingly.
In a similar rowwise-FP8 experiment,
BAPO becomes unstable after approximately 90 steps and eventually collapses
as its clipping bounds move to $[0.9, 3.0]$;
detailed trajectories are provided in Appendix~\ref{sec:bapo-appendix}.
This result indicates that a fixed contribution target does not stabilize this FP8 setting.

\textbf{FP8 RL for LLMs.}
Early efforts on FP8 RL focus exclusively on the rollout stage.
FlashRL \citep{yao2025flashrl} demonstrates that FP8 generation yields an approximately 30\% speedup
but introduces a train–inference mismatch.
Efforts to mitigate this mismatch primarily focus on correcting it via importance sampling.
Truncated Importance Sampling (TIS, \cite{yao2025offpolicy})
and Masked Importance Sampling (MIS, \cite{liu-li-2025-rl-collapse})
provide corrections at the token and sequence levels, respectively.
QuRL \citep{li2026qurl} further refines TIS
by incorporating the proximal-to-behavior policy ratio into the clipping range.

More recent works, including Jet-RL \citep{xi2026jet} and Unified FP8 \citep{lmsys2025fp8rl},
extend FP8 to both rollout and training,
showing that full-pipeline unification narrows the train–rollout precision gap
and facilitates on-policy training.
These works share our motivation but focus on the precision gap;
none identifies the failure mode we characterize:
compounded quantization noise in the importance ratio systematically distorts the clipping mechanism,
causing over-clipping and ultimately destabilizing training.

%% file: conclusion.tex
We investigate the stability of full-pipeline FP8 RL for LLMs and identify a previously overlooked failure mode: compounded FP8 quantization noise distorts the importance ratio, causing systematic over-clipping that triggers pathological entropy surges. To address this, we propose Calibrated Clipping, which dynamically realigns the clipping bounds by matching the FP8 clipping quantile to a high-precision BF16 reference and rebalancing the positive-to-negative update ratio. Extensive experiments demonstrate that Calibrated Clipping consistently eliminates entropy surges and restores BF16-level performance while achieving up to 1.5$\times$ the BF16 training throughput, establishing a practical path toward full-pipeline FP8 RL for LLMs.

%% file: appendix.tex
\subsection{Example of a High-Entropy Response}
\begin{figure}[htbp]
    \centering
    \includegraphics[width=.8\linewidth]{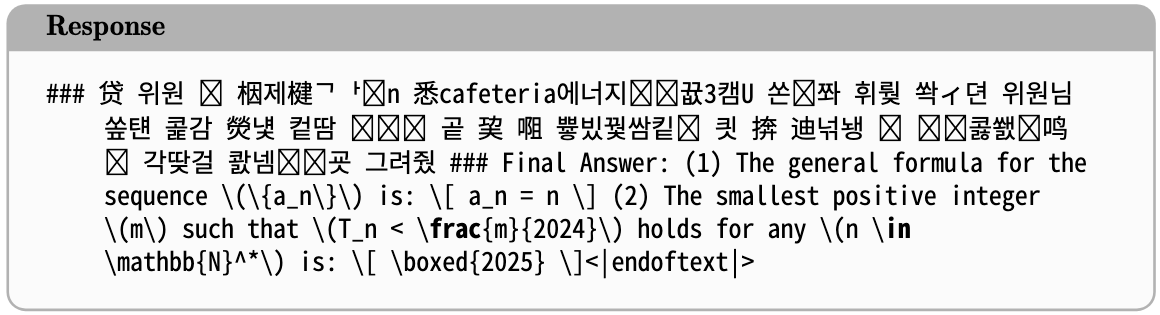}
    \caption{An example of a high-entropy response containing garbled and fragmented tokens.}
    \label{fig:response}
\end{figure}

\subsection{Results of the CISPO Experiment}
\label{sec:cispo-appendix}
\begin{figure}[htbp]
    \centering
    \includegraphics[width=.6\linewidth]{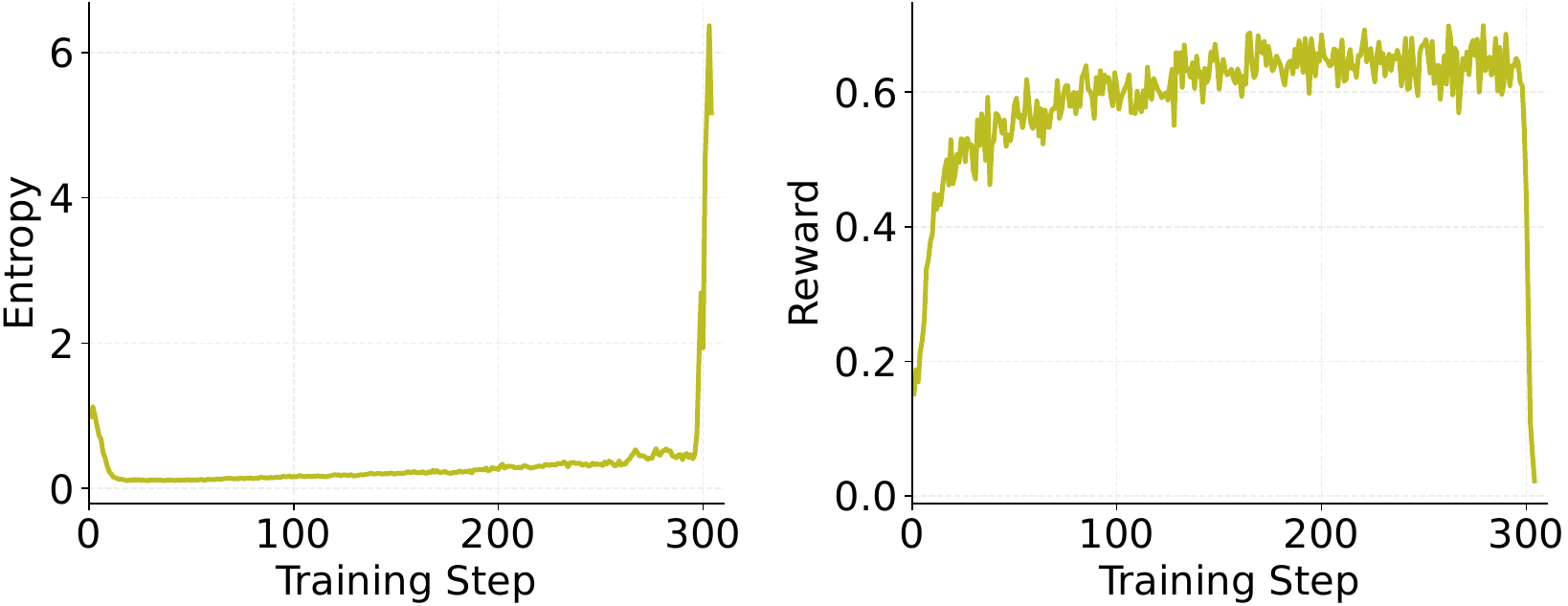}
    \caption{Training entropy and reward of rowwise FP8 CISPO training.}
    \label{fig:cispo}
\end{figure}

We train Qwen3-8B-Base on the DeepScaleR dataset using CISPO, with all remaining configurations identical to those of the GRPO experiments in Section~\ref{sec:grpo-results}. As shown in Figure~\ref{fig:cispo}, training collapses after approximately 300 steps, with entropy spiking sharply and reward dropping to zero. This demonstrates that retaining clipped gradients is insufficient to stabilize training in this rowwise-FP8 setting.

\subsection{Results of the BAPO Experiment}
\label{sec:bapo-appendix}
\begin{figure}[htbp]
    \centering
    \includegraphics[width=\linewidth]{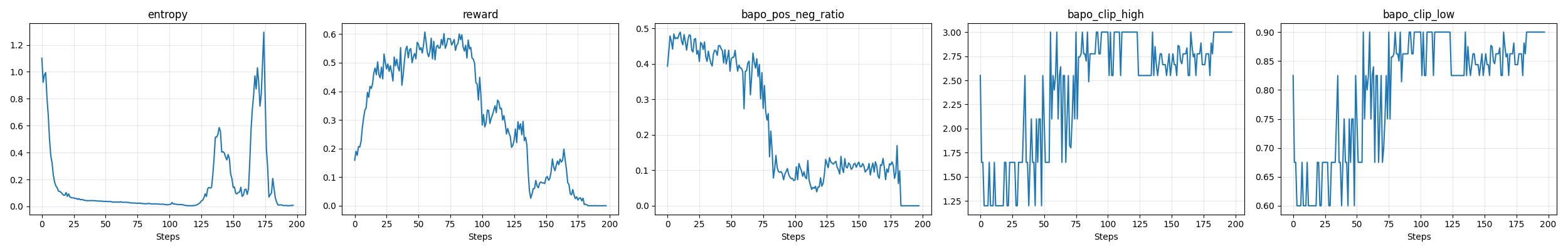}
    \caption{BAPO with rowwise FP8 training in the Qwen3-8B GRPO setting. From left to right: actor entropy, reward, positive-token contribution $\rho$, upper clipping bound, and lower clipping bound.}
    \label{fig:bapo}
\end{figure}

We evaluate BAPO in the same Qwen3-8B GRPO setting as Section~\ref{sec:grpo-results}, using rowwise FP8 training and FP8 rollout. We use the default BAPO configuration \citep{xi2025bapo}: target positive-token contribution $\rho_0=0.4$, lower-bound range $[0.6,0.9]$, upper-bound range $[1.2,3.0]$, and update steps $\delta_{\text{low}}=0.02$ and $\delta_{\text{high}}=0.05$. As shown in Figure~\ref{fig:bapo}, the positive-token contribution $\rho$ falls below its target after approximately 90 steps, while BAPO progressively moves the lower and upper clipping bounds toward 0.9 and 3.0, respectively. Training subsequently becomes unstable, and the reward eventually collapses to zero. In this FP8 setting, using a fixed positive-contribution target therefore fails to stabilize training.

\subsection{Results on the Eurus Dataset}
\label{sec:coding-appendix}
We further evaluate Calibrated Clipping by training Qwen3-8B on the coding subset of Eurus-2-RL-Data \citep{cui2025process} and evaluating on TACO, APPS, and Codeforces. As shown in Table~\ref{tab:coding-results}, Calibrated Clipping improves the average score over vanilla FP8 training for all three scaling granularities. 

\begin{table}[htbp]
\centering
\small
\setlength{\tabcolsep}{5pt}
\renewcommand{\arraystretch}{1.1}
\definecolor{lightgrayrow}{gray}{0.94}
\begin{tabular}{l l c c c c}
\toprule
Rollout & Train & TACO & APPS & Codeforces & Average \\
\midrule
BF16 & BF16 & 37.44 & 52.90 & 58.48 & 49.61 \\
FP8 & BF16 & 35.56 & 54.75 & 55.98 & 48.77 \\
\midrule
FP8 & Tensorwise FP8 & 32.57 & 51.93 & 52.78 & 45.77 \\
\rowcolor{lightgrayrow} FP8 & w/ Calibrated Clipping & 34.83 & 51.64 & 57.15 & 47.88 \\
\midrule
FP8 & Rowwise FP8 & 33.61 & 52.17 & 52.21 & 46.00 \\
\rowcolor{lightgrayrow} FP8 & w/ Calibrated Clipping & 36.44 & 53.45 & 55.83 & 48.58 \\
\midrule
FP8 & Blockwise FP8 & 33.30 & 50.90 & 54.93 & 46.38 \\
\rowcolor{lightgrayrow} FP8 & w/ Calibrated Clipping & 33.61 & 54.35 & 58.26 & 48.75 \\
\bottomrule
\end{tabular}
\caption{Training results of Qwen3-8B under different training precisions and FP8 scaling granularities on Eurus dataset.}
\label{tab:coding-results}
\end{table}

\subsection{Sensitivity to Calibration Hyperparameters}
\label{sec:sensitivity-appendix}
We evaluate the sensitivity of our method to two hyperparameters in the same Qwen3-8B rowwise-FP8 GRPO setting as Section~\ref{sec:case-study}: the recalibration interval and the initialization of the clipping bounds. For the recalibration interval, we compare updates every 10, 20, and 40 training steps. For the initialization, we compare the default bounds $[0.6,1.8]$ with the standard bounds $[0.8,1.2]$.

As shown in Table~\ref{tab:sensitivity-results}, changing the recalibration interval or the initial clipping bounds affects the final average score to some extent, but every tested configuration recovers substantial performance over vanilla rowwise FP8 training, demonstrating the robustness of our method.

\begin{table}[htbp]
\centering
\small
\setlength{\tabcolsep}{5pt}
\renewcommand{\arraystretch}{1.1}
\begin{tabular}{l l c}
\toprule
Ablation & Setting & Average Score \\
\midrule
Update interval & 10 steps & 58.31 \\
Update interval & 20 steps & 56.51 \\
Update interval & 40 steps & 57.91 \\
\midrule
Initialization & $[0.6,1.8]$ & 56.51 \\
Initialization & $[0.8,1.2]$ & 56.20 \\
\bottomrule
\end{tabular}
\caption{Sensitivity of Calibrated Clipping to the recalibration interval and clipping-bound initialization.}
\label{tab:sensitivity-results}
\end{table}

\newpage
\subsection{Experimental Details for GRPO with Qwen3-8B-Base}
\subsubsection{Experiment Settings}
\textbf{Training.} We adopt the DeepScaleR dataset and train with a context length of 16K. The training batch size is 256 with a mini-batch size of 64. We sample eight responses per prompt and train for 500 steps with a learning rate of $10^{-6}$. The TIS truncation hyperparameter is set to $C=2$. We remove the KL-divergence term from the GRPO objective.

\textbf{Evaluation.} We evaluate on all benchmarks using a temperature of 1.0 and a top-$p$ value of 0.7. For AIME24/25, we report Avg@16 accuracy; for all other benchmarks, we report Avg@4 accuracy.

\subsubsection{Training Metrics}
\begin{figure}[H]
    \centering
    \includegraphics[width=0.78\linewidth]{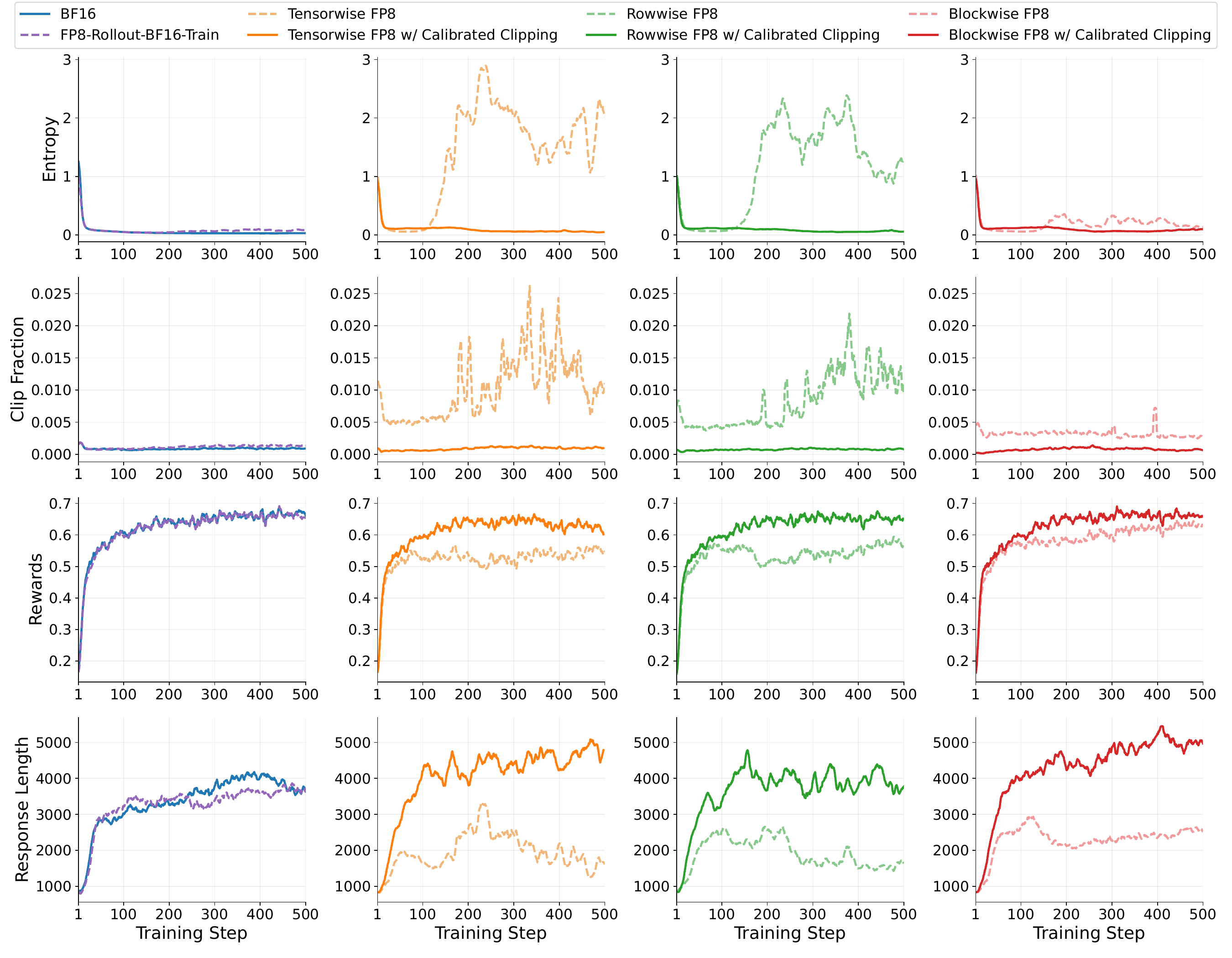}
    \caption{Training results for GRPO experiments with Qwen3-8B-Base.}
    \label{fig:8b_deepsclar}
\end{figure}

\begin{figure}[H]
    \centering
    \includegraphics[width=0.72\linewidth]{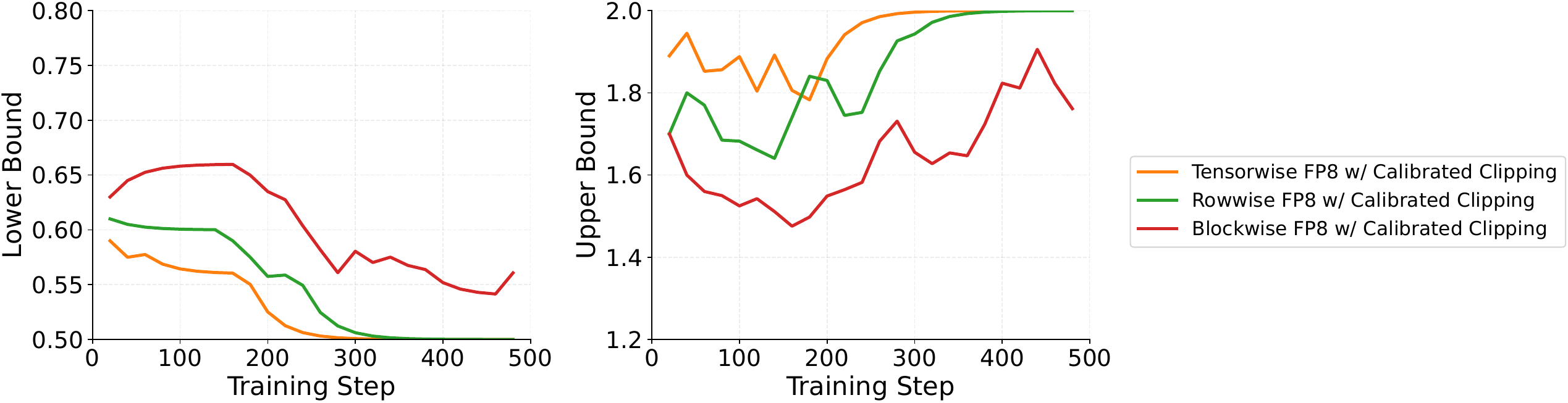}
    \caption{Trajectories of the calibrated lower and upper clipping bounds for GRPO experiments with Qwen3-8B-Base.}
    \label{fig:8b_clip}
\end{figure}

\newpage
\subsection{Experimental Details for GRPO with Qwen2.5-32B}
\subsubsection{Experiment Settings}
All settings are identical to those of the Qwen3-8B-Base experiment. 

\subsubsection{Training Metrics}

\begin{figure}[H]
    \centering
    \includegraphics[width=0.78\linewidth]{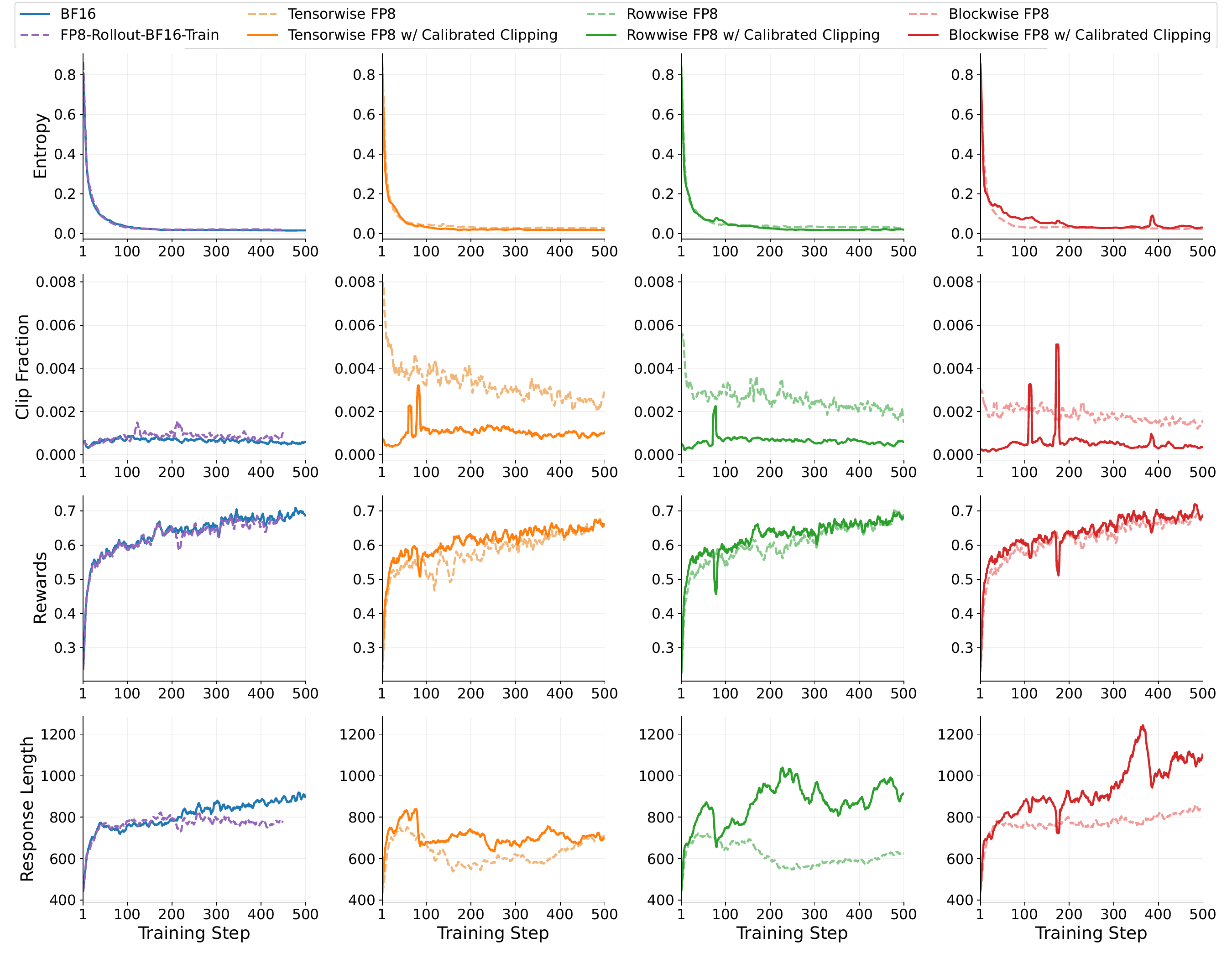}
    \caption{Training results for GRPO experiments with Qwen2.5-32B.}
    \label{fig:32b_deepsclar}
\end{figure}

\begin{figure}[H]
    \centering
    \includegraphics[width=0.72\linewidth]{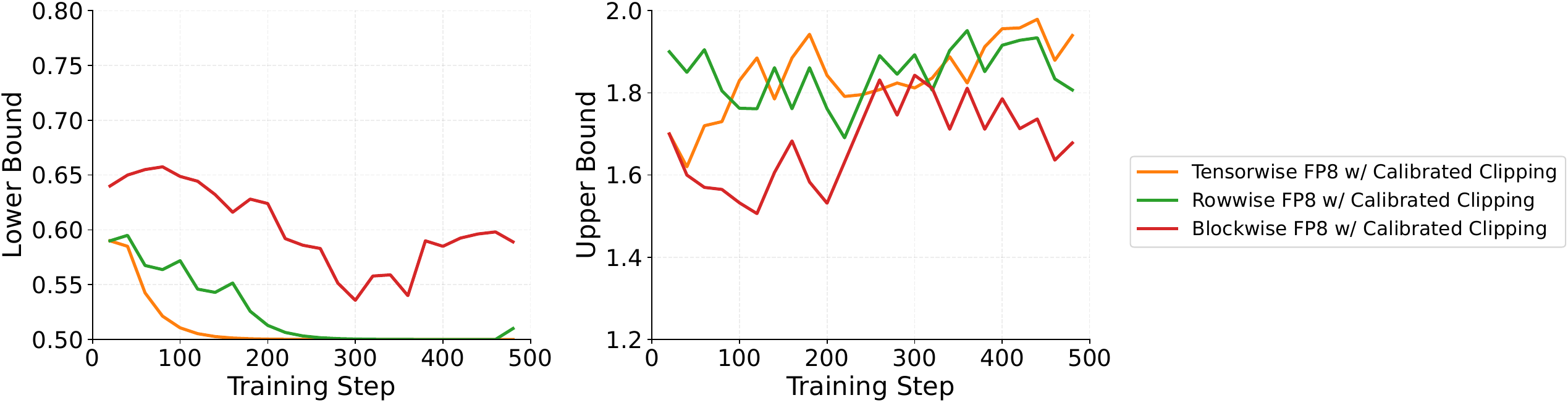}
    \caption{Trajectories of the calibrated lower and upper clipping bounds for GRPO experiments with Qwen2.5-32B.}
    \label{fig:32b_clip}
\end{figure}

\newpage
\subsection{Experimental Details for DAPO with Qwen3-14B-Base}
\subsubsection{Experiment Settings}
\textbf{Training.} We adopt the DAPO-Math-17K dataset and train with a context length of 20K. The training batch size is 256 with a mini-batch size of 64. We sample 16 responses per prompt and train for 200 steps with a learning rate of $10^{-6}$. The TIS truncation hyperparameter is set to $C=2$. We do not apply any KL divergence loss between the actor and reference models. 

\textbf{Evaluation.} We evaluate on the AIME 2024 dataset using a temperature of 1.0 and a top-$p$ value of 0.7 and report Avg@32 accuracy.

\subsubsection{Training Metrics}

\begin{figure}[H]
    \centering
    \includegraphics[width=0.78\linewidth]{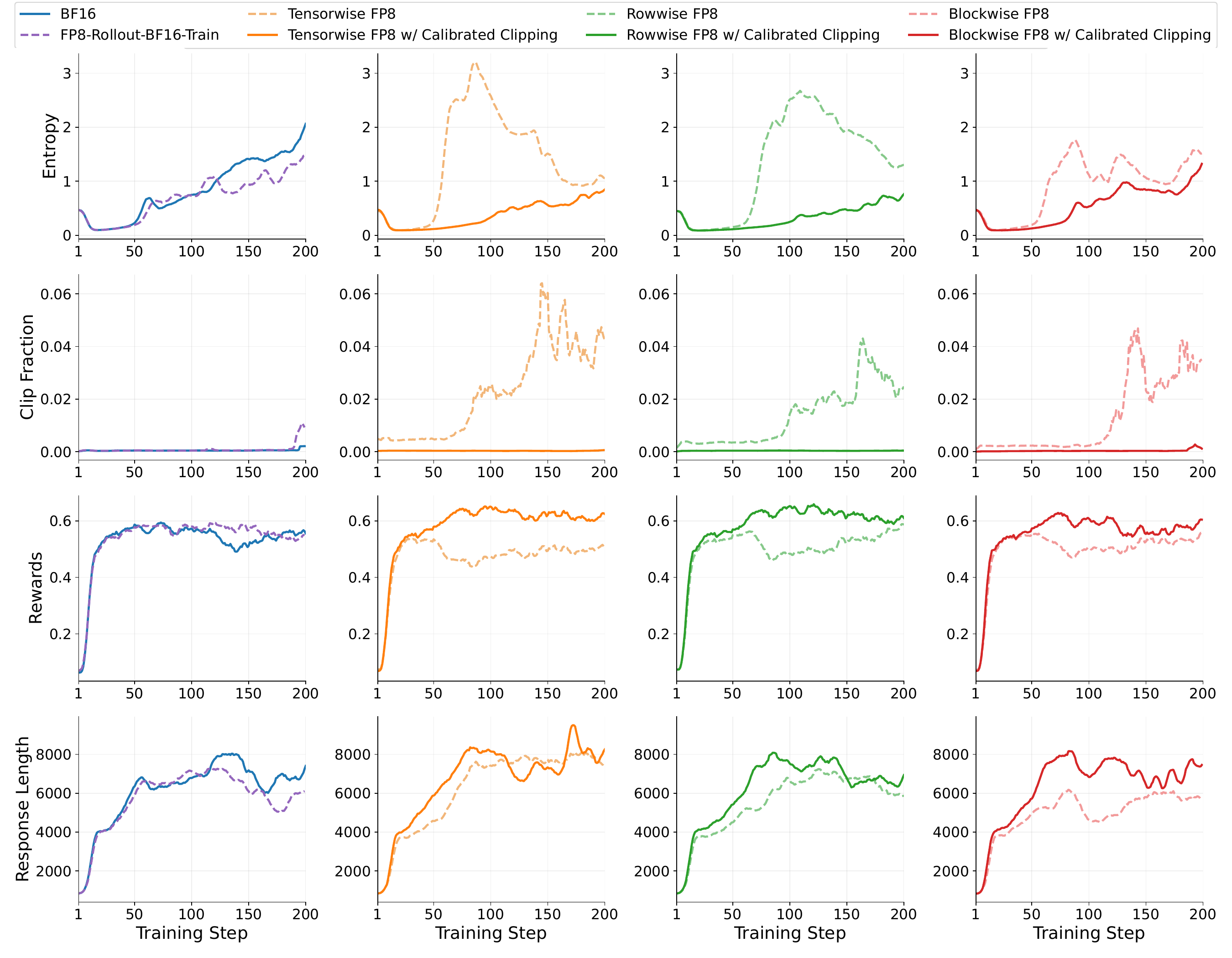}
    \caption{Training results for DAPO experiments with Qwen3-14B-Base.}
    \label{fig:14b_dapo}
\end{figure}

\begin{figure}[H]
    \centering
    \includegraphics[width=0.72\linewidth]{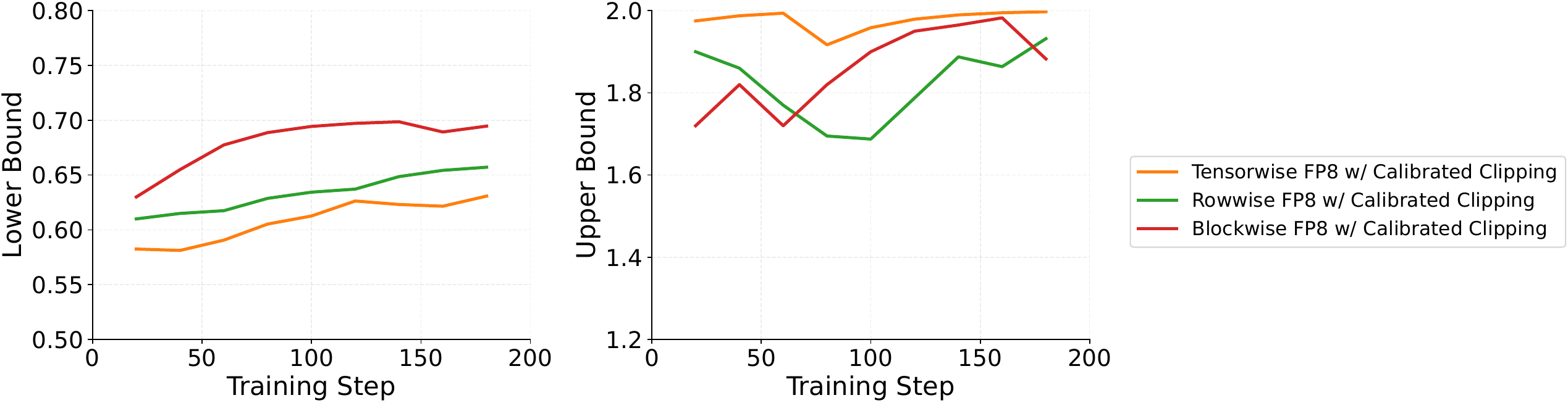}
    \caption{Trajectories of the calibrated lower and upper clipping bounds for DAPO experiments with Qwen3-14B-Base.}
    \label{fig:14b_clip}
\end{figure}